\documentclass[letterpaper, 10 pt, conference, twoside]{ieeeconf}  %
\usepackage{graphicx}
\usepackage{bm}
\usepackage{xcolor}
\usepackage{soul}
\usepackage{amsmath}
\usepackage{hyperref}
\usepackage{amsfonts}
\usepackage{amssymb}
\usepackage{tabularx} 
\usepackage{multicol}
\usepackage{multirow}
\usepackage{siunitx}

\usepackage{courier}

\usepackage{svg}

\usepackage{amssymb}  %
\usepackage{multirow}
\usepackage{url}
\usepackage{breakurl}
\usepackage{enumerate}
\usepackage{subcaption}

\usepackage{listings}

\definecolor{codegreen}{rgb}{0, 0, 1}
\definecolor{codegray}{rgb}{1, 1, 1}
\definecolor{codepurple}{rgb}{1, 1, 1}
\definecolor{backcolour}{rgb}{1, 1, 1}

\lstdefinestyle{mystyle}{
    backgroundcolor=\color{backcolour},   
    commentstyle=\color{codegreen},
    keywordstyle=\color{magenta},
    numberstyle=\tiny\color{codegray},
    stringstyle=\color{codepurple},
    basicstyle=\ttfamily\footnotesize,
    breakatwhitespace=false,         
    breaklines=true,                 
    captionpos=b,                    
    keepspaces=true,                 
    numbers=left,                    
    numbersep=5pt,                  
    showspaces=false,                
    showstringspaces=false,
    showtabs=false,                  
    tabsize=2
}

\newcommand{\bl}[1]{\boldsymbol{#1}}

\IEEEoverridecommandlockouts                              %

\title{\LARGE \bf
RotorPy: A Python-based Multirotor Simulator with Aerodynamics for Education and Research
}

\author{Spencer Folk, James Paulos, and Vijay Kumar %
\thanks{S. Folk and V. Kumar are with the GRASP Laboratory, University of Pennsylvania, Philadelphia, PA, USA, \{\texttt{sfolk, kumar}\}\texttt{@seas.upenn.edu}. J. Paulos is with Treeswift, Philadelphia, PA, USA, \texttt{jpaulos@gmail.com}. This work is not related to Treeswift.}
}

\begin{document}

\maketitle
\thispagestyle{empty}
\pagestyle{empty}

\begin{abstract}
Simulators play a critical role in aerial robotics both in and out of the classroom. 
We present \textit{RotorPy}, a simulation environment written entirely in Python intentionally designed to be a lightweight and accessible tool for robotics students and researchers alike to probe concepts in estimation, planning, and control for aerial robots. 
\textit{RotorPy} simulates the 6-DoF dynamics of a multirotor robot including aerodynamic wrenches, obstacles, actuator dynamics and saturation, realistic sensors, and wind models. 
This work describes the modeling choices for \textit{RotorPy}, benchmark testing against real data, and a case study using the simulator to design and evaluate a model-based wind estimator. 

\end{abstract}

\section{INTRODUCTION}

Dynamics simulation environments aid robotics education and research, providing a playground for rapid experimentation and evaluation of robotic design, perception, and action. 
Aerial robots, or UAVs, are a complicated application domain--unstable dynamics requiring high speed sensors, actuators, controllers, and planners, and complex aerodynamic interactions with the environment and other UAVs--placing added demands on simulation tools necessary for synthesis and analysis.
Existing simulators, driven by target applications, tend to prioritize compute speed with hardware integration like \textit{RotorS} \cite{furrer2016rotors} and \textit{Agilicious} \cite{Foehn22scienceagilicious}, or photorealistic visualization like with \textit{Airsim} \cite{shah2018airsim} and \textit{Flightmare} \cite{song2020flightmare}.
Also, reinforcement learning (RL) for UAVs is sprouting simulation environments like \textit{Gym-PyBullet-Drones} \cite{panerati2021pybullet} purpose-built for use with common Python-based RL toolkits.
The trend seems to be towards increasingly complex and elaborate codebases requiring a high level of expertise to navigate and understand their modeling choices, making it hard to decide whether or not a simulator will fit the needs of a new user. 
To that end, we developed a new simulation environment called \textit{RotorPy}\footnote{github.com/spencerfolk/rotorpy}, which prioritizes accessibility, transparency, and educational value, serving as a tool for learning and exploration in aerial robotics both for students and researchers.
Initially created as a teaching aid for a robotics course at the University of Pennsylvania, \textit{RotorPy} was designed to be lightweight, easy to install, and accessible to engineers with working knowledge of Python. 

This paper introduces \textit{RotorPy}'s modeling choices, structure, and features contributing to an effective environment for probing aspects of UAVs; and then, we present a case study using the simulator to design a model-based wind estimator. 

\begin{figure}
    \centering
    \includegraphics[width=0.7\linewidth]{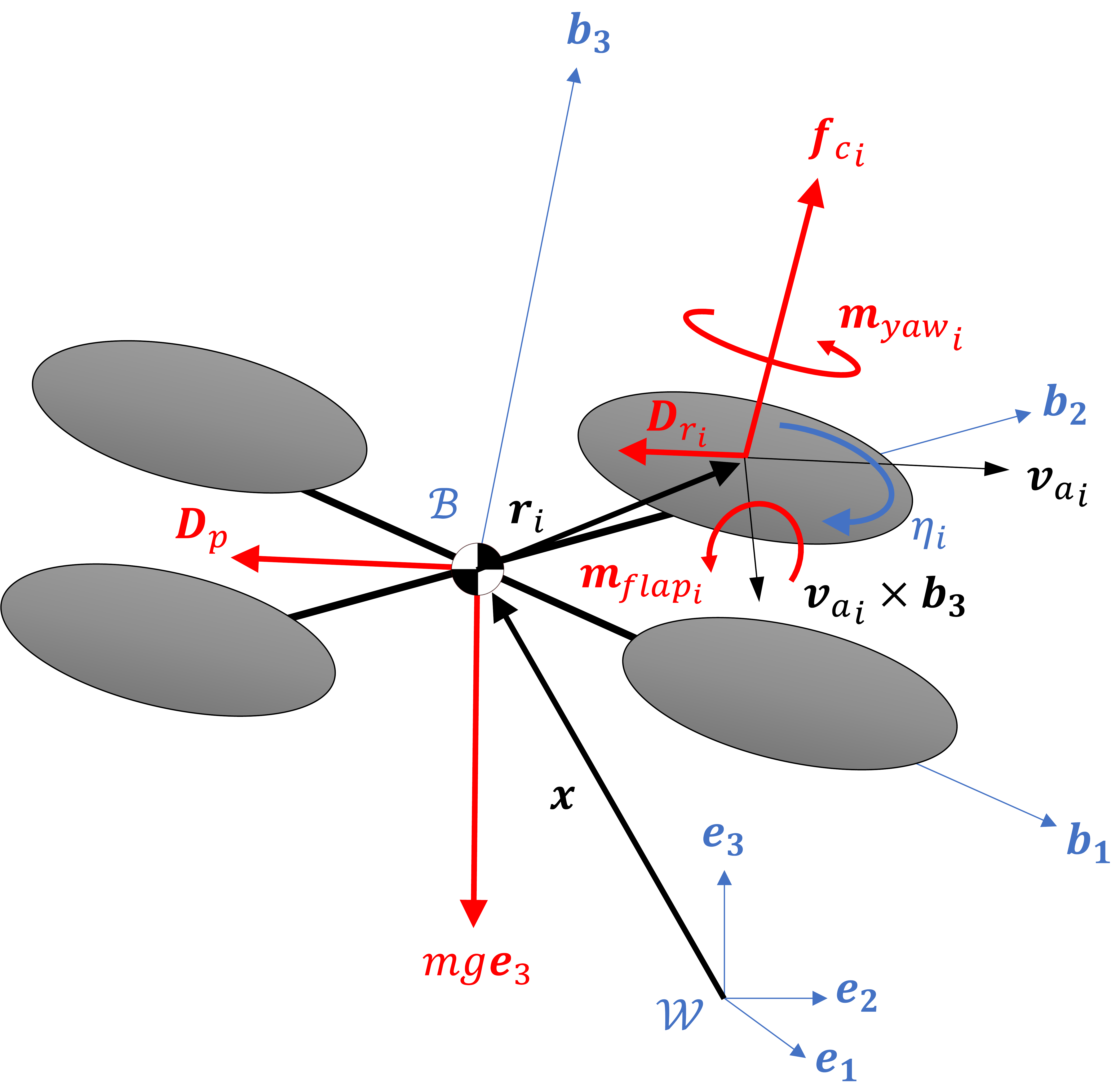}
    \caption{Free body diagram of a UAV subject to control and aerodynamic wrenches. Relative airflow through the fluid medium, $\bl{v}_{a}$, produces additional wrenches in the form of aerodynamic drag on the frame, $\bl{D}_p$ and rotors, $\bl{D}_{r_i}$.}
    \label{fig:quadrotor}
    \vspace{-5mm}
\end{figure}

\section{MODELING}\label{sec:model}

\textit{RotorPy} includes a quadrotor UAV model with aerodynamic wrenches, inertial and motion capture sensors, cuboid obstacle environments, and spatio-temporal wind fields.

\subsection{Multirotor dynamics}
Following Figure \ref{fig:quadrotor}, we model a multirotor UAV with co-planar rotors using the Newton-Euler equations: 
\begin{align}
    \dot{\bl{x}} &= \bl{v} \label{eq:pos_kinematics} \\
    \dot{\bl{v}} &= \frac{1}{m}R(\bl{f}_{c} + \bl{f}_{a})-  g \bl{e}_3 \label{eq:pos_dynamics} \\
    \dot{R} &= R\hat{\bl{\Omega}} \label{eq:att_kinematics} \\
     \dot{\bl{\Omega}} &= J^{-1}( \bl{m}_{c} + \bl{m}_{a} - \bl{\Omega} \times J\bl{\Omega}) \label{eq:att_dynamics} 
\end{align}
where $\bl{x} \in \mathbb{R}^3$ and $\bl{v} \in \mathbb{R}^3$ are respectively the position and velocity vectors; $R \in SO(3)$ is the rotation from the body frame to the world frame; $\bl{\Omega} \in \mathbb{R}^3$ is the angular velocity; $m$ is the total mass and $J$ is the inertia tensor expressed in the body frame. 

The terms $\bl{f}_c \in \mathbb{R}^3$ and $\bl{m}_c \in \mathbb{R}^3$ constitute the control wrench, i.e., the forces and torques produced by the rotor thrust and drag torque. We model the control wrench with
\begin{align}
    \bl{f}_c &= k_\eta \sum_{i=1}^{n}\eta_i^2 \label{eq:f_control} \bl{b}_3 \\
    \bl{m}_c &= k_m \sum_{i=1}^{n} \epsilon_i \eta_i^2 \bl{b}_3 + \sum_{i=1}^n \bl{r}_i \times \bl{f}_{c_i}. \label{eq:m_control}
\end{align}
Here, $\eta_i$ is the $i$'th rotor speed,  $\bl{r}_i$  is the vector from the center of mass to the rotor hub, and $\epsilon_i \in \{-1, 1\}$ is the rotor's direction of rotation. We assume that each rotor has the same static thrust, $k_\eta$, and drag torque, $k_m$, coefficients which can be identified using thrust stand tests.

\subsection{Aerodynamic wrenches} \label{sec:aero}
Together, the vectors $\bl{f}_a \in \mathbb{R}^3$ and $\bl{m}_a \in \mathbb{R}^3$ make up the aerodynamic wrench, which is a collection of forces and torques produced by the relative motion of the UAV through a fluid medium.
There are multiple physical phenomena that produce $\bl{f}_a$ and $\bl{m}_a$ on the UAV (see \cite{svacha2020inertia} and references therein). These effects are all dependent on the relative body airspeed, $\bl{v}_a = R^\top(\dot{\bl{x}} - \bl{w})$ where $\bl{w}\in\mathbb{R}^3$ is a local wind vector in the world frame. We lump these effects into three contributions to the aerodynamic wrench: parasitic drag, rotor drag, and blade flapping. 

\subsubsection{Parasitic drag}
Parasitic drag is the combination of skin friction and pressure drag acting on the body of the UAV. 
It is characteristically proportional to the airspeed squared: 
\begin{equation}
    \label{eq:parasitic}
    \bl{D}_p = -C\vert \vert\bl{v}_a \vert \vert_2\bl{v}_a
\end{equation}
where $C = \text{diag}(c_{Dx}, c_{Dy}, c_{Dz})$ is a matrix of parasitic drag coefficients corresponding to each body axis. 

\subsubsection{Rotor drag}
In contrast to parasitic drag, which is dominant at higher airspeeds, rotor drag can have a surprisingly large presence on small UAVs even at lower airspeeds.
The physical phenomenon responsible for rotor drag is the dissymmetry of lift produced by a rotor in forward flight, whereby the advancing blade experiences a higher airspeed than the retreating blade producing an imbalance of forces on the rotor.
We adopt the rotor drag model used in \cite{svacha2020inertia} in which the drag force is proportional to the product of the airspeed and the rotor speed. 
\begin{equation}
    \label{eq:rotor_drag}
    \bl{D}_{r_i} = -K \eta_i \bl{v}_{a_i} 
\end{equation}
where $K = \text{diag}(k_d, k_d, k_z)$ is a matrix of rotor drag coefficients\footnote{As noted in \cite{svacha2020inertia}, the $k_z$ term isn't actually a source of drag, but rather a linear approximation of loss of thrust due to change in inflow. However, it resembles an effective drag on the body z axis.} corresponding to each body axis. 

\subsubsection{Blade flapping}
Dissymmetry of lift at the advancing and retreating sides of the rotor will also cause the rotor blades to deflect up and down as they revolve in a flapping motion. 
Svacha \textit{et al.} \cite{svacha2020inertia} provides experimental evidence for flapping moments even for small UAVs with rigid rotors. 
This is a very complex phenomenon that can produce both longitudinal and lateral moments depending on the rigidity of the blades \cite{allen1946flapping}. 
Our model expresses blade flapping as a longitudinal moment following \cite{svacha2020inertia} 
\begin{equation}
    \label{eq:flapping_moment}
    \bl{m}_{flap_i} = -k_{flap} \eta_i \bl{v}_{a_i} \times \bl{b}_3
\end{equation} 
with $k_{flap}$ being the flapping coefficient. 

The total aerodynamic force in the body frame is $\bl{f}_a = \bl{D}_p + \sum_{i=1}^n \bl{D}_{r_i}$, and the total moment is $\bl{m}_a = \sum_{i=1}^n (\bl{m}_{flap_i} + \bl{r}_i \times \bl{D}_{r_i})$.

\subsection{Actuator dynamics}

Even for very small UAVs, the motors take time to settle to a commanded speed. 
Capturing this effect has proven to be important especially for RL applications \cite{panerati2021pybullet}. 
We model the actuator delay using a first order process:
\begin{equation} \label{eq:actuator}
    \dot{\bl{\eta}} = \frac{1}{\tau_m}(\bl{\eta}_c - \bl{\eta})
\end{equation}
where $\bl{\eta}_c \in \mathbb{R}^n$ are the commanded rotor speeds and $\tau_m$ is the motor time constant--it can be identified using static thrust stand testing.

\subsection{Sensors}

\subsubsection{Inertial measurement unit}
The simulator's inertial measurement unit (IMU) measurement is given by: 
\begin{equation} \label{eq:imu}
    \bl{h}_{IMU} = \begin{bmatrix} R_\mathcal{I}^\mathcal{B}(\dot{\bl{v}} + g\bl{e}_3) + \bl{a}_{IMU} + \bl{b}_a + \bl{\nu}_a \\ \bl{\Omega} + \bl{b}_g + \bl{\nu}_g\end{bmatrix}
\end{equation}
where $\dot{\bl{v}}$ is given by equation \ref{eq:pos_dynamics}, $\bl{a}_{IMU} = \bl{\Omega}\times(\bl{\Omega}\times\bl{r}_{IMU})$, $\bl{r}_{IMU}$ and $R_\mathcal{I}^\mathcal{B}$ are the position and orientation of the sensor in the body frame, and $\bl{\nu}_{(\cdot)} \sim \mathcal{N}(0, \Sigma_{(\cdot)})$ are sensor noises.
The biases $\bl{b}_{(\cdot)}$ are driven by random walk to simulate drift. 

\subsubsection{External motion capture}

The external motion capture sensor provides information about the pose and twist of the robot in the world frame. 
\begin{equation} \label{eq:mocap}
    \bl{h}_{MC} = \begin{bmatrix} \bl{x} + \bl{\nu}_x \\ \bl{v} + \bl{\nu}_v \\ \bl{q} \oplus \bl{q}_{\nu} \\ \bl{\Omega} + \bl{\nu}_\Omega \end{bmatrix}.
\end{equation}
Above, $\bl{q}$ is the quaternion representation of $R$, $\oplus$ is the quaternion group operation, and $\bl{q}_\nu$ is a quaternion formed by small noise perturbations following \cite{sola2017quaternion}. 

\subsection{Wind}

Wind is modeled by treating the local average wind acting on the center of mass as an additional state vector. 
How this state evolves depends on the chosen wind profile. 
\textit{RotorPy} offers flexibility by supporting both spatial and temporal wind profiles.
Several profiles like step changes, sinusoids, and the Dryden wind turbulence model are included for evaluating controller robustness or estimation accuracy.

\section{SIMULATION FRAMEWORK}

\textit{RotorPy} is written entirely in Python--a deliberate choice originally made to serve instructional needs. 
While this choice might come at a performance cost, the readability and widespread use of Python in scientific computing is the key to this simulator's accessibility and low barrier to entry which is beneficial for education, as originally intended, but also for research. 
Python also makes installation of both \textit{RotorPy} and its dependencies possible with one command.

\subsection{Usage}

\textit{RotorPy} is a collection of modules that can be imported to scripts anywhere. 
The \texttt{Environment} class makes it possible to create, run, and analyze multiple simulations, all with potentially unique configurations, in just a single Python file. 
We believe this design principle makes \textit{RotorPy} stand out among other simulators which typically run in a self-contained manner. 
In contrast, our simulator is purposefully exportable in a manner conducive to studies that require lots of data (e.g., reinforcement learning, design parameter search, controller verification). 

The environment needs a vehicle, controller, and planner; examples of these are all provided out of the box. 
\begin{lstlisting}[language=Python]
sim_instance = Environment(vehicle, controller, trajectory, *args)
\end{lstlisting}
The \texttt{Environment} also has options to add wind and obstacles, configure sensor intrinsics and extrinsics, and more. 

Running the simulator only takes one line: 
\begin{lstlisting}[language=Python]
results = sim_instance.run(duration, *args)
\end{lstlisting}
The output of \texttt{run()} is a dictionary containing the ground truth states, desired state from the trajectory planner, sensor measurements, and controller commands. 
In addition to optional auto-generated plots and simple animations for quick user assessment, we provide a script for automatically converting the results into a \textit{Pandas} DataFrame for larger scale data analysis.

\subsection{Numerical integration}

UAVs are hybrid systems--the dynamics are continuous but control occurs in discrete instances--motivating an approach to numerical integration that preserves the continuity of dynamics in between controller updates. 
To that end, \textit{RotorPy} uses an RK45 integrator with variable step size\footnote{docs.scipy.org/doc/scipy/reference/generated/scipy.integrate.solve\_ivp.html}. 
An added benefit of the variable step size is that we can run simulations with larger time steps, reducing compute cost, while preserving the integration accuracy.

\section{BENCHMARKING}

In order to verify our models, especially for the sensors, we collected flight data from a Crazyflie 2.1\footnote{www.bitcraze.io/} performing a series of aggressive maneuvers. 
In this paper, we present one instance of our hardware trials in which the Crazyflie is commanded to fly in a tight circle at speeds up to 2.5 m/s.   

\begin{figure}[h!]
  \begin{subfigure}[b]{1\columnwidth}
    \includegraphics[width=\linewidth]{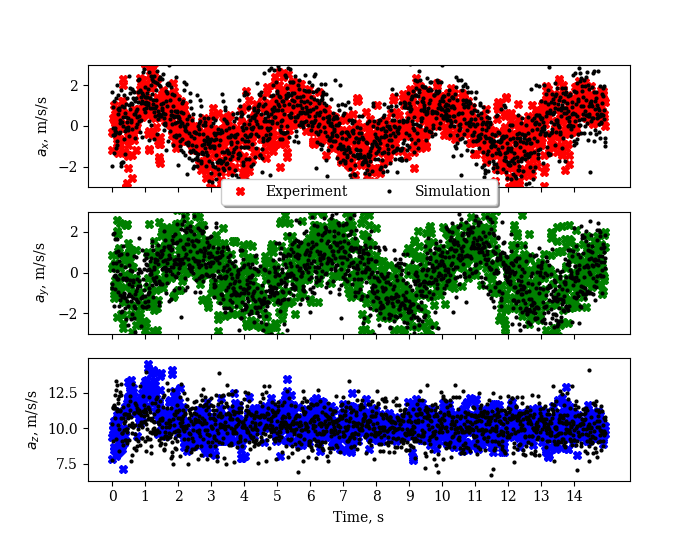}
    \caption{Linear accelerometer measurements in the body frame.}
  \end{subfigure}
  \begin{subfigure}[b]{1\columnwidth}
    \includegraphics[width=\linewidth]{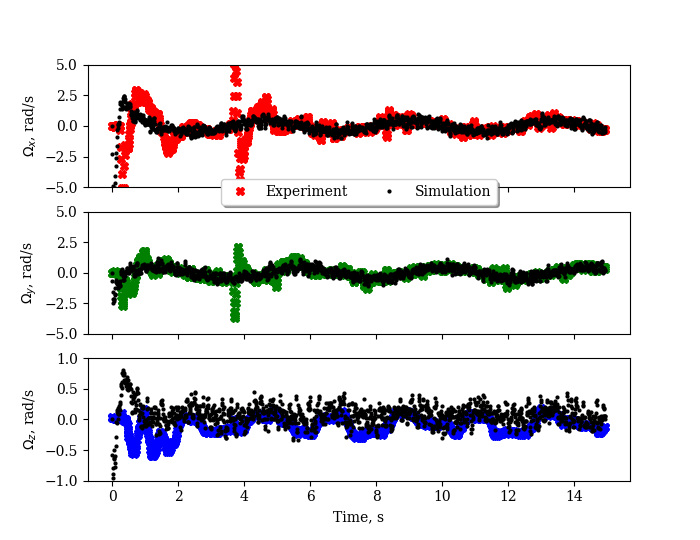}
    \caption{Angular velocity measurements in the body frame.}
  \end{subfigure}
\caption{A comparison between the simulated and actual measurements from the Crazyflie's IMU while tracking a 1.5 m circle at 2.5 m/s.}
\label{fig:circle}
\vspace{-6mm}
\end{figure}

\subsection{Hardware setup}

A motion capture system sends pose and twist data at 100Hz to a base station computer, which uses a nonlinear geometric controller to generate a command based on the current state and desired trajectory. 
\textit{RotorPy}'s controller and trajectory generator are designed to be compatible with our lab hardware. 
The Crazyflie uses onboard PID controllers to track the collective thrust and attitude commands from the simulator's controller.

\subsection{Circle comparison}

In Figure \ref{fig:circle}, we compare the IMU measurements from simulation and the Crazyflie.
This comparison principally highlights the simulator's sensor and aerodynamic models.
For our co-planar configuration, the accelerations in the $x$ and $y$ body axes isolate the drag forces for comparison, since we would otherwise expect zero accelerations in the absence of the aerodynamic wrenches \cite{martin2010feedback}.

\section{CASE STUDY: WIND ESTIMATION}\label{sec:case_study}

We demonstrate the utility of \textit{RotorPy} by evaluating a custom Bayesian filter for estimating the local wind vector, $\bl{w}$, using measurements from the navigation system rather than a dedicated wind sensor--this approach can be classified as indirect wind estimation \cite{abichandani2020wind}.
The estimator is implemented in simulation as an unscented Kalman Filter (UKF), using the simulator's accelerometer and the motion capture sensors to observe the wind vector.
The process model makes several simplifications: linearized attitude dynamics and a version of the aerodynamics that only considers parasitic drag. 
For each evaluation, a calibration procedure collects simulated flight data of the quadrotor with randomized parameters, and then fits quadratic drag coefficients for the process model. 

\begin{figure}[h!]
    \centering
    \includegraphics[width=\columnwidth]{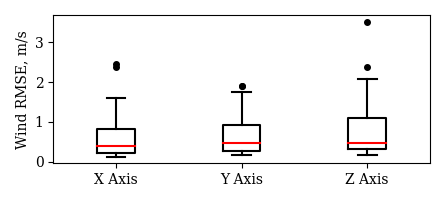}
    \caption{Monte Carlo evaluation of the wind filter over 50 simulations; each instance has randomized mass, drag coefficients, and average wind magnitudes.}
    \label{fig:windmonte}
    \vspace{-2mm}
\end{figure}

\begin{figure}[h]
    \centering
    \includegraphics[width=\columnwidth]{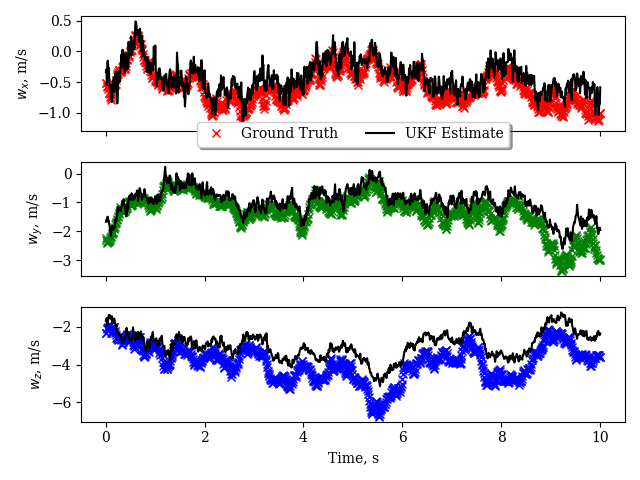}
    \caption{A simulated instance of the unscented Kalman Filter estimating the local wind velocity vector for a quadrotor subject to Dryden wind gusts.}
    \label{fig:windest}
    \vspace{-2mm}
\end{figure}

Figure \ref{fig:windmonte} summarizes the average RMSE over 50 evaluations of the filter using randomized quadrotor parameters in Table \ref{tb:monte}.
In half of the trials, the filter's RMSE falls around or under 0.5 m/s; however, performance is poor in cases where the calibration procedure fails to find good drag coefficients, like when the real drag coefficients are small.
Figure \ref{fig:windest} is one instance of the evaluation, comparing the actual wind components to that estimated from the filter. 
This trial highlights an important model discrepancy: the process model uses the commanded thrust, not the actual thrust, produced by the rotors.
In cases of overwhelming winds like in Figure \ref{fig:windest}, the motors are saturated which causes a model discrepancy between the commanded and actual thrust, leading to estimation error. 

\begin{table}[h!]
\caption{Randomized quadrotor parameters for the Monte Carlo evaluation of the wind filter. Symbols are consistent with Section \ref{sec:model}.}
\label{tb:monte}
\begin{center}
\begin{tabular}{ c | c c }
 \textbf{Parameter} & \textbf{Unit} & \textbf{Range} \textbf{(min-max)} \\ 
 \hline
 $m$  & kg & 0.375--0.9375 \\  
 $c_{Dx}$ & N-(m/s)$^{-2}$ & 0--1($10^{-3}$)  \\
 $c_{Dy}$ & N-(m/s)$^{-2}$ & 0--1($10^{-3}$) \\
 $c_{Dz}$ & N-(m/s)$^{-2}$ & 0--2($10^{-2}$) \\
 $k_d$ & N-rad-m-s$^{-2}$ & 0--1.19($10^{-3}$)\\
 $k_z$ & N-rad-m-s$^{-2}$ & 0--2.32($10^{-3}$)
\end{tabular}
\end{center}
\vspace{-5mm}
\end{table}

\section{SUMMARY}

This work presents \textit{RotorPy}, a UAV simulation environment that is designed to be accessible to engineers with working knowledge of Python. 
The simulator is packaged with a 6-DoF model of a quadrotor UAV with aerodynamics and motor dynamics, realistic sensors, obstacles, and wind models. 
In addition, we provide a tracking controller, multiple trajectory generation methods, and a wind estimation filter for convenience.
For verification, we compare our simulator to real data collected from a Crazyflie performing aggressive trajectories that highlight the aerodynamic forces present in high speed flight.
We believe that \textit{RotorPy} can be a useful tool both in and out of the classroom as a way to dig deep into concepts in estimation, planning, and control for UAVs in the presence of high winds. 
This is demonstrated in our case study, which looks at using \textit{RotorPy} to evaluate a model-based wind estimator.
Future developments include broader support for different UAV archetypes and incorporation of a fast fluid dynamics solver for native spatio-temporal wind field generation.

\addtolength{\textheight}{-12cm}   %

\bibliographystyle{IEEEtran}
\bibliography{refs}

\end{document}